# Subjective Sentiment Analysis for Arabic Newswire Comments

Sadik Bessou, Rania Aberkane
Department of Computer Science, Faculty of Sciences
University of Ferhat Abbas Sétif 1, Algeria
{bessou.s@univ-setif.dz} {aberkan.rania93@gmail.com}

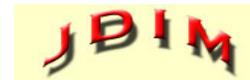

**ABSTRACT:** *This paper presents an approach based on supervised machine learning methods to discriminate between positive, negative and neutral Arabic reviews in online newswire. The corpus is labeled for subjectivity and sentiment analysis (SSA) at the sentence-level. The model uses both count and TF-IDF representations and apply six machine learning algorithms; Multinomial Naïve Bayes, Support Vector Machines (SVM), Random Forest, Logistic Regression, Multi-layer perceptron and k-nearest neighbors using uni-grams, bi-grams features. With the goal of extracting users' sentiment from written text.*

*Experimental results showed that n-gram features could substantially improve performance; and showed that the Multinomial Naïve Bayes approach is the most accurate in predicting topic polarity. Best results were achieved using count vectors trained by combination of word-based uni-gramsand bi-grams with an overall accuracy of 85.57% over two classes and 65.64% over three classes.*

**Review Metrics:** Review Scale- 0/6, Review Score-4.85, Inter-reviewer Consistency- 82%

**DOI:** 10.6025/jdim/2019/17/5/289-295

## 1. Introduction

With the explosion of communication technologies and the accompanying pervasive use of social media, we notice an outstanding proliferation of reviews, comments, recommendations and other forms of opinion expressions. This opinionated content attracted researchers from different fields; economy, political sciences, social sciences, psychology and particularly language processing. One of the prominent subjects is the sentiment analysis also called opinion mining. Sentiment analysis (SA) is the process of identifying and extracting subjective information, sentiments, opinions in a text using natural language processing and machine learning techniques. The problem is usually addressed by formulating the SA as a classification task. The task involves identifying whether a text expresses a Positive, a Negative, or a Neutral sentiment. Applications of SA are varied including analyzing social media output to survey the public, to gain an overview of the wider public opinions and emotions towards certain persons, topics, products or services, to predict stock market, to build personalized recommendation systems, to track the public mood, etc. These applications can be applied in different fields for instance: economy, business, education politics, sports, tourism, etc., which helps in the decision-making. However, SA is still far from producing perfect results due to the complexity of the language. The use of Arabic language has been increasing consistently over various social media platforms. However, Arabic imposes many challenges due to its complex morphology and agglutinative nature with a highly inflectional and derivational system, Arabic words have different polarity categories in different contexts and users use frequently dialectal Arabic (DA) rather than modern standard Arabic (MSA).

In this work, we provide a new resource to support research advances in Arabic sentiment analysis (ASA). We scrap comments from an Algerian online newspaper (Echorouk online)[1]. We highlight some features that discriminate between the different sentiment polarities. We propose a supervised approach, which relies on training language models for the collected data to discriminate comment sentiments based on *n*-gram-words using six machine-learning algorithms.

In the remainder of this paper, we review related work in section 2 and report on data and methods in section 3; where we present datasets, and we describe our approach. Finally, we discuss results and analyze errors in section 4.

## 2. Related Work

The objective of this section is to provide a review of the major works that have been devoted to ASA. A number of projects were conducted and several studies published applied on both MSA and DA.

Some researchers addressed the problem of building SA resources. Abdul-Mageed & Diab (2012) [2] presented the AWATIF multi-genre corpus of MSA labeled for subjectivity and sentiment analysis (SSA) at the sentence-level. The corpus was labeled using both manually and crowd sourcing annotation using Penn Arabic Treebank, Wikipedia Talk pages, and web forums. In another study Abdul-Mageed & Diab (2014) [3] presented SANA a large scale multi-genre, multi dialect lexicon for the SSA of MSA, Egyptian and Levantine dialects. SANA is developed both manually and automatically exploiting data from several genres like Arabic Treebank newswire, Twitter, YouTube comments, and Egyptian chat logs.

Aly & Atiya (2013) [6] presented LABR a large scale Arabic book reviews dataset, where the reviews are rated on a scale of 1 to 5 stars. These data were used for both rating classification and sentiment polarity classification.

ElSahar & El-Beltagy (2015) [13] build several domain specific datasets for ASA. The domains covered in the dataset were movies, hotels, restaurants and products. The lexicon was extracted using a semi-supervised approach.

Nabil et al. (2015) [20] presented ASTD (Arabic social sentiment analysis dataset) gathered from Twitter, the tweets are classified as objective, subjective positive, subjective negative, and subjective mixed. Authors performed two sets of benchmark experiments: four-way sentiment classification and two-stage classification.

Other researchers tackled the problem of classification, Shoukry & Rafea (2012) [22] considered a corpus-based approach for SA of tweets written in MSA and Egyptian dialects. They collected 1000 tweets divided equally into positive and negative. After filtering the tweets, they used standard n-gram features and experimented with SVM and Naïve Bayes classifiers.

Ibrahim et al. (2015) [15] presented a SA system for MSA and Egyptian dialect using a corpus of different types of data. Authors used rich feature sets to improve the classification by handling the valence shifters, question and supplication terms. The experimental results showed good performance with SVM classifier.

Mourad & Darwish (2013) [19] improved the accuracy of SSA of Arabic tweets by translating an English lexicon, and applying a random graph walk approach on a manually created lexicon using Arabic-English AMT phrase tables. The authors added different features such as stemming, POS tagging, tweet-specific features, etc.

Khalifa & Omar (2014) [16] presented a hybrid approach for Arabic opinion question answering and applied it to Arabic costumers reviews on Jordanian hotels. The approach consists of extracting named entities and determining the polarities of words in the reviews. The authors experimented with Naïve Bayes, SVM and KNN classifiers.

Tartir & Abdul-Nabi (2017) [23] presented a semantic approach to detect user attitudes and business insights in social media using MSA or DA. The twitter feeds are classified using Arabic sentiment ontology (ASO) into positive or negative. The approach produced good understanding.

El-Masri et al. (2017) [12] presented a web based tool that applied SA to Arabic text tweets. Several parameters are proposed: the time of the tweets, preprocessing, n-gram features, lexicon-based methods, machine-learning methods. The polarity labels are (positive, negative, both, and neutral). Experimental results showed that Naïve Bayes approach performs better than other classifiers.

Alomari et al. (2017) [5] investigated different supervised machine learning SA approaches applied to Arabic user's social media. Authors constructed their corpus by collecting Arabic tweets written in Jordanian dialect and MSA. Experiments were conducted using SVM and Naïve Bayes algorithms utilizing different features and preprocessing strategies.

Heikal et al. (2018) [14] explored different deep learning models to predict the sentiment of Arabic tweets. They used an ensemble model, combining convolutional neural network (CNN) and long short-Term Memory (LSTM) models. The model achieved significant improvement in the F1-score and the accuracy over the existing models.

---

[1] https://www.echoroukonline.com/

Abdul-Mageed (2018) [1] introduced a framework of structural and social context features of the Twitter domain and showed its utility in classification with an SVM approach.

Baly et al. (2019) [8] created: the Multi-Dialect Arabic Sentiment Twitter Dataset, and the Arabic Sentiment Twitter Dataset for the Levantine dialect. The authors experimented with SVM, logistic regression and random forest trees classifiers using POS tags, numbers of positive/negative emoticons and words from different lexicons, Twitter-specific features, etc., in addition to several deep learning models.

The survey of Al-Ayyoub et al. (2019) [4] presented a noteworthy comprehensive overview of the works done on ASA. The survey grouped 361 published papers based on the SA related problems. It covered the methods, tools, and resourced that used in the ASA. The aspects considered in the study were Binary/ternary SSA, Multi-Way SA, Aspect-Based SA, Multilingual SA, and Other SA-related problems. The study covered both corpus-based and lexicon-based SA approaches and it covered dialects and MSA.

Recently, several evaluation campaigns were dedicated for SSA. SemEval is the international workshop on semantic evaluation. It is an ongoing series of evaluations of computational semantic analysis systems. It has been run yearly since 2013 [21]. The first time SemEval introduced Arabic language for all subtasks, was in 2017. [21]

Our proposed approach focuses on word-based n-gram language models using machine-learning algorithms expecting significant improvement in accuracy.

## 3. Data and Methods

### 3.1 Dataset
There are many available standard datasets for English sentiment classification, but unfortunately, for Arabic there is no standard dataset. Almost of the researchers collect their own corpus from the online web sites. Consequently, no common dataset is used for benchmarking results and evaluating experiments. [10]

The major impact of using online data sources rather than standard datasets is the kind of data. Reviews and comments are opinionated and include a subjective information rather than descriptive one.

Our dataset consists of Arabic comments and reviews that are manually harvested from online newspaper. We use "Echorouk online", a daily newspaper in Algeria; it supports reviews and comments allowing readers to express their opinions about the article they are reading. It is the most read, and it is the second most visited website in 2018 in Algeria.[2]

We collect different articles in economy, politics, social issues, violence, culture, art, etc. Our corpus as shown in table 1 consists of 1.633 documents with 63.055 tokens. Each comment has been annotated for sentiment polarity: positive, negative, neutral: 31.392 tokens for negative, 21.248 tokens for neutral and 9.975 tokens for positive. The annotation is manually conducted to guarantee the best results.

To collect this corpus we search articles with a number of opinionated comments superior than 20.Welabel each comment with one of the following tags: positive, negative, neutral.

| Sentiment label | # documents | # tokens |
|---|---|---|
| Positive | 453 | 9.975 |
| Negative | 760 | 31.392 |
| Neutral | 420 | 21.248 |
| Total | 1.633 | 63.055 |

Table 1. Number of documents and tokens in each label

### 3.2 Preprocessing
The scrapped data necessitate pre-processing since noisy and worthless information data can decrease the efficiency of the system. To improve the quality of the input data we clean up the unwanted content by performing the following preprocessing steps:

**Removal of URLs**
Comments contain frequent web links to share additional information. The content of the links is not analyzed, hence the link itself does not provide any useful information and its removal can reduce the feature size.

**Filtering**
The purpose of filtering is to remove character sequences that may be noisy and thus affect the quality of data. After converting text corpus into UTF"8 encoding, it is necessary to clean up the texts by removing punctuation marks, special characters, non-Arabic characters, dates, time, numbers, single letters, links, and diacritics, etc. None of these impurities represents any polarity. Therefore, they should be removed.

**Tokenizing**
It consists on splitting paragraphs into sentences and sentences into tokens or words. In this step, we normalize our data based on white space, excluding all non-Unicode characters.

---

[2] https://www.alexa.com/topsites/countries/DZ

**Normalizing**

Normalization means replacing specific letters within the word with other letters according to a predefined set of rules; i.e., the unification of characters. Some writing forms (Hamza and Alif) need normalization, which consists for instance in converting "إ", "أ" and "آ" into "ا" because most of the Arabic texts neglect the addition of Hamza on Alif. Another kind of impurity encountered is the elongation where users repeat letters for exaggeration. We shorten the elongated words by replacing the repeated letters with a simple occurrence instead.

**Stop Words Removal**

Stop words (pronouns, conjunctions, prepositions, and names) are extremely frequent words and considered as valueless for taking them as features. We remove stop words that do not affect the classification task. Negation words should not be removed; they reverse the sentiment from positive to negative and vice versa.

**Stemming**

It is the process of removing affixes from words, and reducing these words to their roots. It can significantly improve the efficiency of the classification by reducing the number of terms being input to the classification [9].

Many stemming methods have been developed for Arabic language. The two most widely used stemming methods are:

1. The heavy stemming: Allows transforming each surface Arabic word in the document into its root. [17]

2. The light stemming: Allows removing prefixes and suffixes. [18]

In this work, we use light stemming. That does not reduce a word to its proper root but it removes only prefixes and suffixes from words, as the removal of infixes can change the word meaning completely and consequently the sentiment polarity.

**3.3 Training and Test Datasets**

In this step, we label the comments, whose total number is 1.633 comments with 63.055 tokens, and then we split the dataset into training set and test set. The training set consists of 1.306 comments (80%), and the test set of 327comments (20%).

**3.4 Proposed Model**

Machine learning algorithms can predict sentiments based on textual data. Sentiment analysis based on machine learning consists of classifying subjective texts in two or more categories. The binary classification determines whether the text expresses a positive or a negative opinion. A multi-way classification determines whether the text expresses a positive, a negative or a neutral opinion. Data needs to be annotated with sentiment labels. Labelled data are fed into the machine learning algorithm to build a classification model, which in turn can predict the label for unforeseen instances. Naive Bayes and SVM are commonly used for sentiment classification, with satisfactory results [7], they have showed good accuracy in sentiment polarity classification in various languages [11], such as English, Chinese and Arabic.

Our approach is focused on word-based *n*-grams using various classification algorithms, since syntactic units and relations are expressed at the word-level. We extract different lengths of *n*-grams, 1-2 word *n*-grams. These *n*-grams are used as features in the vector space model (VSM) which builds a term-document matrix by assigning a weight to every term appears in each comment. Many schemes of this model can be used. In our case, we use Count vectors based on combinations of 1-2 word *n*-grams (binary weights) and term-frequency inverse-document-frequency (TF-IDF) vectors based on combinations of 1-2 word *n*-grams (sophisticated weights). For each sentiment polarity, we train a word-level language model.

We formulate the task as a multi-class classification problem, where each sentiment polarity is a separate class. Given a collection of comments and associated polarities, we consider a supervised system to predict the sentiment labels of the comments, $f: C \rightarrow P_i$. It assigns to each comment $C$, the sentiment polarity $P_i$ that maximizes its conditional probability score $argmax_i P(P_i \backslash C)$. For this task, we use six algorithms: Multinomial Naïve Bayes (MNB), Random Forest (RF), Linear SVC (LSVC), K-Nearest Neighbors (KNN), Logistic regression (LR) and Multi-layer perceptron (MLP).

The goal of the experiments is to find the highest accuracy using different classifiers.

We used default settings for Logistic Regression and Multinomial Naïve Bayes. For Linear Support Vector Machine we changed the number of iterations to 1500.

In the Logistic Regression and the Linear Support Vector Machine, we used L1 and L2 regularization, which can be added to the algorithm to ensure that the models do not over fit their data. The L1 regularization norm is the sum of the absolute differences between the estimated and target values, while the L2 regularization norm is the sum of square of the differences between estimated and target values. The regularization value of 1.0 have been used for class weighting. For Multinomial Naïve Bayes we used Laplace smoothing regularization method.

**4. Results and Discussion**

In this section, we illustrate our conducted experiments. We have performed two different experiments by training six classifiers using diverse choices of features. We explore the problem of sentiment classification as a two-class classification problem: "positive, negative" and a three-class classification problem: "positive, negative and neutral". The feature set is composed of *uni*-grams and *bi*- grams represented with count vectors. We tested Bag-of-words features using *uni*-gram, *bi*-grams, and both. The results of the different classifiers using all set of features are compared to determine which classifier is most accurate in the task. Results are reported in term of accuracy, precision, recall, and F1-score. We notice that using count vector weighting scheme through different *n*-grams gives much better results than TF-IDF weighting scheme.

### 4.1 Experiments over Two Classes
In this set of experiments, we apply a binary classification that determines whether the comment expresses a positive or a negative opinion. Therefore, we disregard the neutral class.

**First Experiment**
The first experiment is conducted on the model trained using *uni*-grams features. The results of the evaluation are shown in table 2.

| Algorithm | Accuracy | Precision | Recall | F1-score |
|---|---|---|---|---|
| MNB | **84.71%** | **85%** | **85%** | **84%** |
| RF | 75.61% | 76% | 76% | 76% |
| LSVC | 72.31% | 74% | 72% | 73% |
| KNN | 43.80% | 72% | 44% | 35% |
| LR | 77.68% | 78% | 78% | 78% |
| MLP | 79.33% | 79% | 79% | 79% |

Table 2. Results of experiments over two classes using *uni*-grams

Multinomial Naïve Bayes achieved the best accuracy with 84.71 % and the best results for the other metrics.

**Second Experiment**
The following experiments are performed using *bi*-grams. The results are presented in Table 3.

| Algorithm | Accuracy | Precision | Recall | F1-score |
|---|---|---|---|---|
| MNB | **75.20%** | **76%** | **75%** | **73%** |
| RF | 51.23% | 71% | 51% | 48% |
| LSVC | 51.75% | 72% | 52% | 48% |
| KNN | 37.19% | 64% | 37% | 22% |
| LR | 74.79% | 76% | 75% | 72% |
| MLP | 55.75% | 75% | 56% | 54% |

Table 3. Results of experiments over two classes using *bi*-grams

Multinomial Naïve Bayes achieved the best accuracy with 75.20 % and the best results for the other metrics.

**Third Experiment**
This experiment represents the accuracy of previous algorithms with combination of *uni*-grams and *bi*-grams features. The results are shown in table 4.

| Algorithm | Accuracy | Precision | Recall | F1-score |
|---|---|---|---|---|
| MNB | **85.57%** | **86%** | **86%** | **85%** |
| RF | 70.66% | 73% | 71% | 71% |
| LSVC | 70.66% | 74% | 71% | 72% |
| KNN | 40.49% | 69% | 40% | 29% |
| LR | 79.75% | 80% | 80% | 80% |
| MLP | 83.05% | 83% | 83% | 83% |

Table 4. Results of experiments over two classes using both *uni*-grams and *bi*-grams

Multinomial Naïve Bayes achieved the best accuracy with 85.57% and the best results for the other metrics. The accuracy when combining *uni*-grams and *bi*-grams is the best over the three experiments.

### 4.2 Discussion
Given the experimental results, we notice a low performance when dealing with *bi*-grams alone. The combination of *uni*-grams and *bi*-grams outperformed the use of *bi*-grams by around 10% of accuracy reporting 85.57%. Furthermore, we notice that Multinomial Naïve Bayes classifier performs better than the other classifiers, and the results are improved when combining *uni*-grams and *bi*-grams. The best recall and precision are achieved by Multinomial Naïve Bayes classifier at 86% and 86%. Multinomial Naïve Bayes is followed by Multi-layer Perceptron, which achieves high performance results outperforming the remaining classifiers with an accuracy of 83.05%. While considerable improvements are gained for all classifiers, only slight performance in accuracy (40.49%) is reached for KNN.

### 4.3 Experiments over Three Classes
In this set of experiments, all dataset instances with the three-class labels are used.

**Fourth experiment**
The fourth experiment is conducted on the model that is trained using *uni*-grams features. The results of the evaluation are presented in table 5.

We notice that Multinomial Naïve Bayes is the best classifier. It achieves the highest accuracy with 64.41%.

**Fifth experiment**
The fifth experiment is performed using *bi*-grams features. The results are shown in Table 6.

In this experiment, there is a significant reduction in accuracy for all classifiers. We can find out that Multinomial

| Algorithm | Accuracy | Precision | Recall | F1-score |
|---|---|---|---|---|
| MNB | **64.41%** | 65% | 64% | 63% |
| RF | 62.57% | 63% | 63% | 58% |
| LSVC | 60.73% | 59% | 61% | 59% |
| KNN | 37.11% | 40% | 37% | 30% |
| LR | 62.88% | 61% | 63% | 61% |
| MLP | 63.20% | 61% | 63% | 61% |

Table 5. Results of experiments over three classes using *uni*-grams

Naïve Bayes and Logistic Regression have the same accuracy: 58.58%. However, when observing F1-score Multinomial Naïve Bayes achieves the highest result. Therefore, it is the best classifier.

**Sixth experiment**
This experiment represents the accuracy of previous algorithms combing *uni*-grams and *bi*-grams features. The results are shown in table 7.

| Algorithm | Accuracy | Precision | Recall | F1-score |
|---|---|---|---|---|
| MNB | **65.64%** | 67% | 66% | 64% |
| RF | 60.42% | 64% | 60% | 58% |
| LSVC | 61.34% | 61% | 61% | 59% |
| KNN | 31.90% | 42% | 32% | 23% |
| LR | 62.88% | 61% | 62% | 59% |
| MLP | 63.19% | 68% | 67% | 63% |

Table 7. Results of experiments over three classes using both *uni*-grams and *bi*-grams

We notice that Multinomial Naïve Bayes achieves the highest accuracy of 65.64%, comparing to the other classifiers and comparing to the two previous experiments. The accuracy when combining *uni*-grams and *bi*-grams is the best over the three experiments.

**4.4 Discussion**
As expected, the introduction of the neutral class causes a reduction in accuracy. The sentiment classification into three classes is more difficult than two-class classification. Multinomial Naïve Bayes proved to be the best performing classifier scoring a significant difference than the rest of classifiers reporting 65.64% of accuracy.

We notice a low performance when dealing with *bi*-grams alone. The combination of *uni*-grams and *bi*-grams outperformed the use of *bi*-grams by around 7% of accuracy. The best recall and precision are achieved by Multi-layer Perceptron classifier at 68% and 67%. Multinomial Naïve Bayes is followed by Multi-layer Perceptron, which achieves high performance results outperforming the remaining classifiers with an accuracy of 63.20%. While considerable improvements are gained for all classifiers, only slight performance in accuracy (31.90%) is reached for KNN.

Over all experiments, we found out that preprocessing, *n*-grams combination, and count vectors representation weighting improve the classification performance. In these experiments, six supervised machine-learning classifiers were compared for sentiment classification. The experimental results show that Multinomial Naïve Bayes outperformed the other classifiers. We conclude that it is better to combine *uni*-grams and *bi*-grams to improve performance on the sentiment classification.

**5. Conclusion**

In this paper, we used machine learning to detect sentiments in online newswire comments. Several models were trained for sentence-level SA. Our model relies on a pre-trained word vector representation. We used various classifiers, features, and preprocessing strategies to find the best models to predict the sentiment label. The results showed that *n*-gram features could substantially improve performance. Additionally, we noticed that the kind of data representation could provide a significant performance boost compared to simple representation. The best performing feature representation is the combination of *uni*-grams and *bi*-grams.

Out of the experimental results, we highlighted that the best performing classifier was Multinomial Naïve Bayes and the worst was KNN. The findings show that although subjectivity and sentiment expressed at semantic and pragmatic levels modeling them can benefit from lower linguistics levels in lexical space.

In the future, we plan to extend our work to investigate more complex emotion recognition models and explore dialectal Arabic as well as experiment on multi-genre, multi-lingual lexical resources, and multi-level (sentence, paragraph, and document), and investigate more algorithms, which may be insightful. In light of the recent successes of deep learning models, we plan to experiment similarly with deep learning techniques on the task.

**References**

[1] Abdul-Mageed, M. (2018). Learning Subjective Language: Feature Engineered vs. Deep Models. *In*: *LRECWorkshop The 3rd Workshop on Open-Source Arabic Corpora and Processing Tools*. 80-90.

[2] Abdul-Mageed, M., Diab, M. T. (2012). AWATIF: A Multi-Genre Corpus for Modern Standard Arabic Subjectivity and Sentiment Analysis. *In*: LREC, 3907-3914.

[3] Abdul-Mageed, M., Diab, M. T. (2014). SANA: A Large

Scale Multi-Genre, Multi-Dialect Lexicon for Arabic Subjectivity and Sentiment Analysis. *In*: LREC, 1162-1169.

[4] Al-Ayyoub, M., Khamaiseh, A. A., Jararweh, Y., Al-Kabi, M. N. (2019). A comprehensive survey of arabic sentiment analysis. *Information Processing & Management*, 56(2) 320-342.

[5] Alomari, K. M., ElSherif, H. M., Shaalan, K. (2017). Arabic Tweets Sentimental Analysis Using Machine Learning. *In*: *International Conference on Industrial, Engineering and Other Applications of Applied Intelligent Systems*. 602-610. Springer, Cham.

[6] Aly, M., Atiya, A. (2013). LABR: A large scale Arabic book reviews dataset. *In*: Proceedings of the 51st Annual Meeting of the Association for Computational Linguistics (Volume 2: Short Papers). 2, 494-498.

[7] Assiri, A., Emam, A., Aldossari, H. (2015). Arabic sentiment analysis: a survey. *International Journal of Advanced Computer Science and Applications*, 6(12) 75-85.

[8] Baly, R., Khaddaj, A., Hajj, H., El-Hajj, W., Shaban, K. B. (2019). ArSentD-LEV: A multi-topic corpus for target-based sentiment analysis in Arabic levantine tweets. arXiv preprint arXiv:1906.01830.

[9] Bessou, S., Touahria, M. (2014). An Accuracy-Enhanced Stemming Algorithm for Arabic Information Retrieval. *Neural Network World*, 24(2) 117-128.

[10] Boudad, N., Faizi, R., Thami, R. O. H., Chiheb, R. (2017). Sentiment analysis in Arabic: A review of the literature. *Ain Shams Engineering Journal*.

[11] Elarnaoty, M., AbdelRahman, S., Fahmy, A. (2012). A machine learning approach for opinion holder extraction in Arabic language. arXiv preprint arXiv:1206.1011.

[12] El-Masri, M., Altrabsheh, N., Mansour, H., Ramsay, A. (2017). A web-based tool for Arabic sentiment analysis. *Procedia Computer Science*, 117, 38-45.

[13] ElSahar, H., El-Beltagy, S. R. (2015). Building large Arabic multi-domain resources for sentiment analysis. *In*: *International Conference on Intelligent Text Processing and Computational Linguistics*. 23-34. Springer, Cham.

[14] Heikal, M., Torki, M., El-Makky, N. (2018). Sentiment Analysis of Arabic Tweets using Deep Learning. *Procedia Computer Science*, 142, 114-122.

[15] Ibrahim, H. S., Abdou, S. M., Gheith, M. (2015). Sentiment analysis for modern standard Arabic and colloquial. arXiv preprint arXiv:1505.03105.

[16] Khalifa, K., Omar, N. (2014). A Hybrid method using Lexicon-based Approach and Naive Bayes Classifier for Arabic Opinion Question Answering. JCS, 10(10) 1961-1968.

[17] Khoja, S., Garside, R. (1999). Stemming Arabic text. Lancaster, UK, Computing Department, Lancaster University.

[18] Larkey, L. S., Connell, M. E. (2001). Arabic information retrieval at UMass in TREC-10. *In*: TREC.

[19] Mourad, A., Darwish, K. (2013). Subjectivity and sentiment analysis of modern standard Arabic and Arabic microblogs. *In*: Proceedings of the 4th workshop on computational approaches to subjectivity, sentiment and social media analysis (p. 55-64).

[20] Nabil, M., Aly, M., Atiya, A. (2015). ASTD: Arabic sentiment tweets dataset. *In*: Proceedings of the 2015 Conference on Empirical Methods in Natural Language Processing, 2515-2519.

[21] Rosenthal, S., Farra, N., Nakov, P. (2017). SemEval-2017 task 4: Sentiment analysis in Twitter. *In*: Proceedings of the 11th International Workshop on Semantic Evaluation (SemEval-2017) 502-518.

[22] Shoukry, A., Rafea, A. (2012). Sentence-level Arabic sentiment analysis. *In*: 2012 International Conference on Collaboration Technologies and Systems (CTS) (p. 546-550). IEEE.

[23] Tartir, S., Abdul-Nabi, I. (2017). Semantic sentiment analysis in Arabic social media. *Journal of King Saud University-Computer and Information Sciences*, 29(2) 229-233.